\def\year{2021}\relax
\documentclass[letterpaper]{article} 
\usepackage{aaai21}  
\usepackage{times}  
\usepackage{helvet} 
\usepackage{courier}  
\usepackage[hyphens]{url}  
\usepackage{graphicx} 
\usepackage{natbib}  
\usepackage{caption} 
\usepackage{multirow}
\usepackage{array}
\usepackage{booktabs}
\usepackage{amsmath}
\usepackage{amssymb}
\usepackage{dsfont}

\newcommand{\Ttr}{T_\text{train}}
\newcommand{\Tte}{T_\text{test}}

\title{Graph Neural Network-Based Anomaly Detection in Multivariate Time Series}
\author {
    Ailin Deng, Bryan Hooi \\
}
\affiliations {
    National University of Singapore \\
    ailin@comp.nus.edu.sg, bhooi@comp.nus.edu.sg
}
 
\begin{document}

\maketitle

\begin{abstract}

Given high-dimensional time series data (e.g., sensor data), how can we detect anomalous events, such as system faults and attacks? More challengingly, how can we do this in a way that captures complex inter-sensor relationships, and detects and explains anomalies which deviate from these relationships? Recently, deep learning approaches have enabled improvements in anomaly detection in high-dimensional datasets; however, existing methods do not explicitly learn the structure of existing relationships between variables, or use them to predict the expected behavior of time series. Our approach combines a structure learning approach with graph neural networks, additionally using attention weights to provide explainability for the detected anomalies. Experiments on two real-world sensor datasets with ground truth anomalies show that our method detects anomalies more accurately than baseline approaches, accurately captures correlations between sensors, and allows users to deduce the root cause of a detected anomaly.

\end{abstract}

\section{Introduction}

With the rapid growth in interconnected devices and sensors in Cyber-Physical Systems (CPS) such as vehicles, industrial systems and data centres, there is an increasing need to monitor these devices to secure them against attacks. This is particularly the case for critical infrastructures such as power grids, water treatment plants, transportation, and communication networks. 

Many such real-world systems involve large numbers of interconnected sensors which generate substantial amounts of time series data. For instance, in a water treatment plant, there can be numerous sensors measuring water level, flow rates, water quality, valve status, and so on, in each of their many components. Data from these sensors can be related in complex, nonlinear ways: for example, opening a valve results in changes in pressure and flow rate, leading to further changes as automated mechanisms respond to the change.

As the complexity and dimensionality of such sensor data grow, humans are increasingly less able to manually monitor this data. This necessitates automated anomaly detection approaches which can rapidly detect anomalies in high-dimensional data, and explain them to human operators to allow them to diagnose and respond to the anomaly as quickly as possible.

Due to the inherent lack of labeled anomalies in historical data, and the unpredictable and highly varied nature of anomalies, the anomaly detection problem is typically treated as an unsupervised learning problem. In past years, many classical unsupervised approaches have been developed, including linear model-based approaches~\cite{pca}, distance-based methods~\cite{knn}, and one-class methods based on support vector machines~\cite{oneclasssvm}. However, such approaches generally model inter-relationships between sensors in relatively simple ways: for example, capturing only linear relationships, which is insufficient for complex, highly nonlinear relationships in many real-world settings. 

Recently, deep learning-based techniques have enabled improvements in anomaly detection in high-dimensional datasets. For instance, Autoencoders (AE)~\cite{autoencoding} are a popular approach for anomaly detection which uses reconstruction error as an outlier score. More recently, Generative Adversarial Networks (GANs)~\cite{li2019mad} and LSTM-based approaches~\cite{lstmencoder} have also reported promising performance for multivariate anomaly detection. However, most methods do not explicitly learn which sensors are related to one another, thus facing difficulties in modelling sensor data with many potential inter-relationships. This limits their ability to detect and explain deviations from such relationships when anomalous events occur.

How do we take full advantage of the complex relationships between sensors in multivariate time series? Recently, graph neural networks (GNNs)~\cite{gcnn} have shown success in modelling graph-structured data. These include graph convolution networks (GCNs)~\cite{gcn}, graph attention networks (GATs)~\cite{gat} and multi-relational approaches~\cite{relationalgcn}. However, applying them to time series anomaly detection requires overcoming two main challenges. Firstly, different sensors have very different behaviors: e.g. one may measure water pressure, while another measures flow rate. However, typical GNNs use the same model parameters to model the behavior of each node. Secondly, in our setting, the graph edges (i.e. relationships between sensors) are initially unknown, and have to be learned along with our model, while GNNs typically treat the graph as an input.

Hence, in this work, we propose our novel Graph Deviation Network (\textsc{GDN}) approach, which learns a graph of relationships between sensors, and detects deviations from these patterns\footnote{The code is available at https://github.com/d-ailin/GDN}. Our method involves four main components: 1) \textbf{Sensor Embedding}, which uses embedding vectors to flexibly capture the unique characteristics of each sensor; 2) \textbf{Graph Structure Learning} learns the relationships between pairs of sensors, and encodes them as edges in a graph; 3) \textbf{Graph Attention-Based Forecasting} learns to predict the future behavior of a sensor based on an attention function over its neighboring sensors in the graph; 4) \textbf{Graph Deviation Scoring} identifies and explains deviations from the learned sensor relationships in the graph. 


To summarize, the main contributions of our work are:
\begin{itemize}
\item We propose \textsc{GDN}, a novel attention-based graph neural network approach which learns a graph of the dependence relationships between sensors, and identifies and explains deviations from these relationships. 
\item We conduct experiments on two water treatment plant datasets with ground truth anomalies. Our results demonstrate that \textsc{GDN} detects anomalies more accurately than baseline approaches.
\item We show using case studies that \textsc{GDN} provides an explainable model through its embeddings and its learned graph. We show that it helps to explain an anomaly, based on the subgraph over which a deviation is detected, attention weights, and by comparing the predicted and actual behavior on these sensors.
\end{itemize}

\section{Related Work}

We first review methods for anomaly detection, and methods for multivariate time series data, including graph-based approaches. Since our approach relies on graph neural networks, we summarize related work in this topic as well.

\paragraph{Anomaly Detection} 

Anomaly detection aims to detect unusual samples which deviate from the majority of the data. 
Classical methods include density-based approaches~\cite{breunig2000lof}, linear-model based approaches~\cite{pca}, distance-based methods~\cite{knn}, classification models~\cite{oneclasssvm}, detector ensembles~\cite{featurebagging} and many others. 

More recently, deep learning methods have achieved improvements in anomaly detection in high-dimensional datasets. These include approaches such as autoencoders (AE)~\cite{autoencoding}, which use reconstruction error as an anomaly score, and related variants such as variational autoencoders (VAEs)~\cite{vae}, which develop a probabilistic approach, and autoencoders combining with Gaussian mixture modelling~\cite{dagmm}. 

However, our goal is to develop specific approaches for multivariate time series data, explicitly capturing the graph of relationships between sensors.

\paragraph{Multivariate Time Series Modelling}  
These approaches generally model the behavior of a multivariate time series based on its past behavior. A comprehensive summary is given in \cite{survey_timeseries}.


Classical methods include auto-regressive models~\cite{hautamaki2004outlier} and the auto-regressive integrated moving average (ARIMA) models~\cite{arima_1, arima_2}, based on a linear model given the past values of the series. However, their linearity makes them unable to model complex nonlinear characteristics in time series, which we are interested in.



To learn representations for nonlinear high-dimensional time series and predict time series data, deep learning-based time series methods have attracted interest. These techniques, such as Convolutional Neural Network (CNN) based models~\cite{cnn_deepant}, Long Short Term Memory (LSTM)~\cite{lstm_1,lstm_2, lstmvae} and Generative Adversarial Networks (GAN) models~\cite{zhou2019beatgan,li2019mad}, have found success in practical time series tasks. However, they do not explicitly learn the relationships between different time series, which are meaningful for anomaly detection: for example, they can be used to diagnose anomalies by identifying deviations from these relationships.

Graph-based methods provide a way to model the relationships between sensors by representing the inter-dependencies with edges. Such methods include probabilistic graphical models, which encode joint probability distributions, as described in~\cite{pgm_1,pgm_2}. However, most existing methods are designed to handle stationary time series, and have difficulty modelling more complex and highly non-stationary time series arising from sensor settings.



\paragraph{Graph Neural Networks}
In recent years, graph neural networks (GNNs) have emerged as successful approaches for modelling complex patterns in graph-structured data.
In general, GNNs assume that the state of a node is influenced by the states of its neighbors. Graph Convolution Networks (GCNs)~\cite{gcn} model a node's feature representation by aggregating the representations of its one-step neighbors. Building on this approach, graph attention networks (GATs)~\cite{gat} use an attention function to compute different weights for different neighbors during this aggregation. 
Related variants have shown success in time-dependent problems: for example, GNN-based models can perform well in traffic prediction tasks~\cite{traffic_2,traffic_1}. Applications in recommendation systems~\cite{lim2020stp,relationalgcn} and relative applications ~\cite{wang2020detecting} verify the effectiveness of GNN to model large-scale multi-relational data.

However, these approaches use the same model parameters to model the behavior of each node, and hence face limitations in representing very different behaviors of different sensors. Moreover, GNNs typically require the graph structure as an input, whereas the graph structure is initially unknown in our setting, and needs to be learned from data.



\begin{figure}[t]
\centering
\includegraphics[width=.5\textwidth]{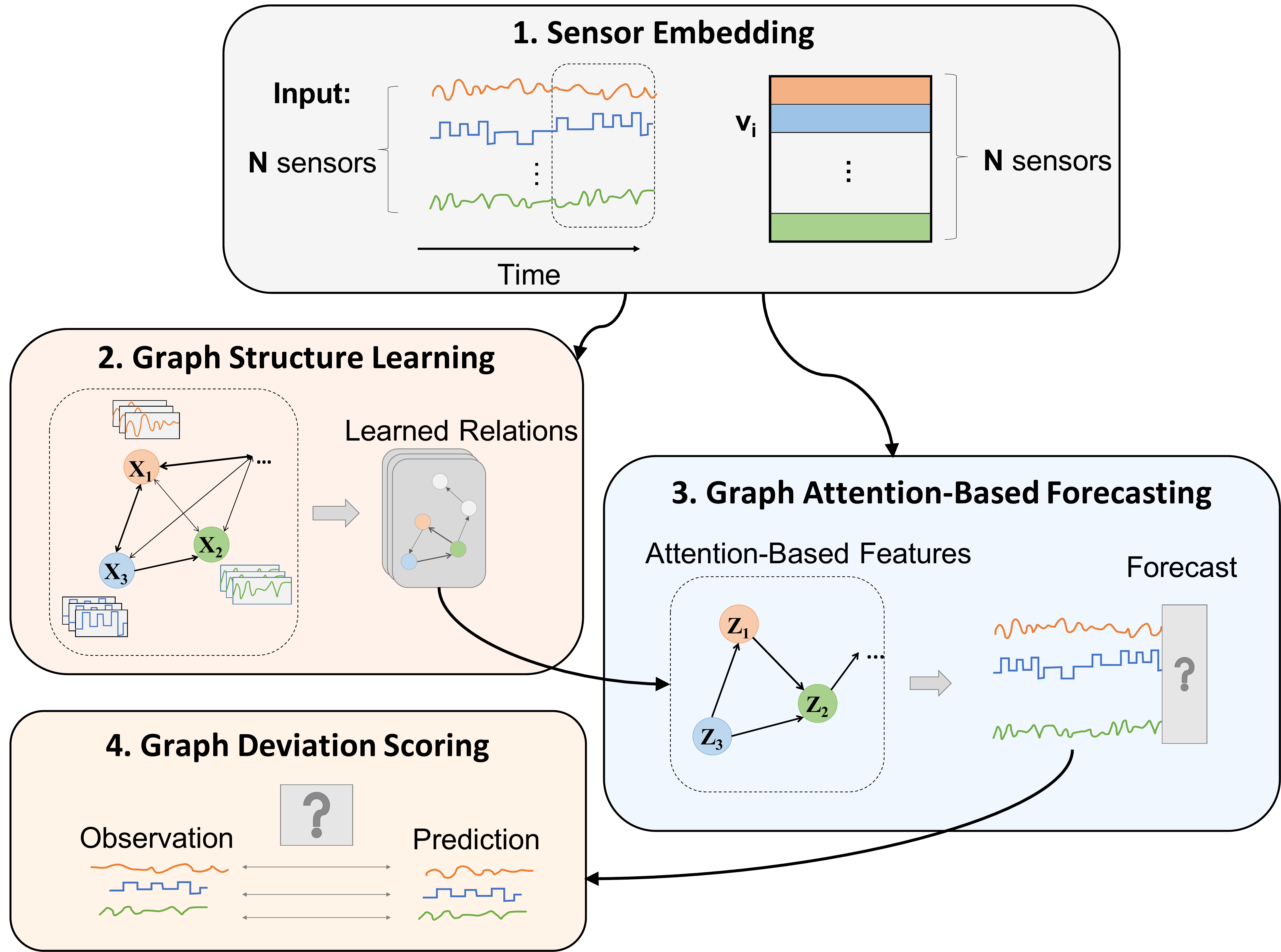}
\caption{Overview of our proposed framework.}

\label{fig:framework}
\end{figure}

\section{Proposed Framework}

\subsection{Problem Statement}
In this paper, our training data consists of sensor (i.e. multivariate time series) data from $N$ sensors over $\Ttr$ time ticks: the sensor data is denoted $\mathbf{s}_\text{train} = \left[\mathbf{s}^{(1)}_\text{train}, \cdots,\mathbf{s}^{(\Ttr)}_\text{train}\right]$, which is used to train our approach. In each time tick $t$, the sensor values $\mathbf{s}^{(t)}_\text{train} \in \mathbb{R}^N$ form an $N$ dimensional vector representing the values of our $N$ sensors. Following the usual unsupervised anomaly detection formulation, the training data is assumed to consist of only normal data.

Our goal is to detect anomalies in testing data, which comes from the same $N$ sensors but over a separate set of $\Tte$ time ticks: the test data is denoted $\mathbf{s}_\text{test} = \left[\mathbf{s}^{(1)}_\text{test},\cdots,\mathbf{s}^{(\Tte)}_\text{test}\right]$. 

The output of our algorithm is a set of $\Tte$ binary labels indicating whether each test time tick is an anomaly or not, i.e. $\mathsf{a}(t) \in \{0, 1\}$, where $\mathsf{a}(t) = 1$ indicates that time $t$ is anomalous. 

\subsection{Overview}
Our \textsc{GDN} method aims to learn relationships between sensors as a graph, and then identifies and explains deviations from the learned patterns. It involves four main components: 
\begin{enumerate}
    \item \textbf{Sensor Embedding}: uses embedding vectors to capture the unique characteristics of each sensor;
    \item \textbf{Graph Structure Learning}: learns a graph structure representing dependence relationships between sensors;
    \item \textbf{Graph Attention-Based Forecasting}: forecasts future values of each sensor based on a graph attention function over its neighbors;
    \item \textbf{Graph Deviation Scoring}: identifies deviations from the learned relationships, and localizes and explains these deviations.
\end{enumerate}
Figure \ref{fig:framework} provides an overview of our framework.



\subsection{Sensor Embedding}
In many sensor data settings, different sensors can have very different characteristics, and these characteristics can be related in complex ways. For example, imagine we have two water tanks, each containing a sensor measuring the water level in the tank, and a sensor measuring the water quality in the tank. Then, it is plausible that the two water level sensors would behave similarly, and the two water quality sensors would behave similarly. However, it is equally plausible that sensors within the same tank would exhibit strong correlations. Hence, ideally, we would want to represent each sensor in a flexible way that captures the different `factors' underlying its behavior in a multidimensional way. 

Hence, we do this by introducing an \textbf{embedding vector} for each sensor, representing its characteristics: 
\begin{align*}
    \mathbf{v_i} \in \mathbb{R}^d,\text{ for }i \in \{1, 2, \cdots, N\}
\end{align*}
These embeddings are initialized randomly and then trained along with the rest of the model.

Similarity between these embeddings $\mathbf{v_i}$ indicates similarity of behaviors: hence, sensors with similar embedding values should have a high tendency to be related to one another. In our model, these embeddings will be used in two ways: 1) for structure learning, to determine which sensors are related to one another, and 2) in our attention mechanism, to perform attention over neighbors in a way that allows heterogeneous effects for different types of sensors.  

\subsection{Graph Structure Learning}

A major goal of our framework is to learn the relationships between sensors in the form of a graph structure. To do this, we will use a \textbf{directed graph}, whose nodes represent sensors, and whose edges represent dependency relationships between them. An edge from one sensor to another indicates that the first sensor is used for modelling the behavior of the second sensor. We use a directed graph because the dependency patterns between sensors need not be symmetric. We use an adjacency matrix $A$ to represent this directed graph, where $A_{ij}$ represents the presence of a directed edge from node $i$ to node $j$.

We design a flexible framework which can be applied either to 1) the usual case where we have no prior information about the graph structure, or 2) the case where we have some prior information about which edges are plausible (e.g. the sensor system may be divided into parts, where sensors in different parts have minimal interaction). 

This prior information can be flexibly represented as a set of \textbf{candidate relations} $\mathcal{C}_i$ for each sensor $i$, i.e. the sensors it could be dependent on:
\begin{align}
    \mathcal{C}_i \subseteq \{1, 2, \cdots, N\} \setminus \{i\}
\end{align}
In the case without prior information, the candidate relations of sensor $i$ is simply all sensors, other than itself.



To select the dependencies of sensor $i$ among these candidates, we compute the similarity between node $i$'s embedding vector, and the embeddings of its candidates $j \in \mathcal{C}_i$:
\begin{align}
    e_{ji}  & = \frac{\mathbf{v_i}^\top \mathbf{v_j}}{\left\|\mathbf{v_i}\right\| \cdot \left\|\mathbf{v_j}\right\|} \text{ for }j \in \mathcal{C}_i\\ 
    A_{ji}  & = \mathds{1}\{ j \in \mathsf{TopK}(\{e_{ki}: k \in \mathcal{C}_i\}) \}
\end{align}
That is, we first compute $e_{ji}$, the normalized dot product between the embedding vectors of sensor $i$, and the candidate relation $j \in \mathcal{C}_i$. Then, we select the top $k$ such normalized dot products: here $\mathsf{TopK}$ denotes the indices of top-$k$ values among its input (i.e. the normalized dot products). The value of $k$ can be chosen by the user according to the desired sparsity level. Next, we will define our graph attention-based model which makes use of this learned adjacency matrix $A$.

\subsection{Graph Attention-Based Forecasting}

In order to provide useful explanations for anomalies, we would like our model to tell us:
\begin{itemize}
    \item Which sensors are deviating from normal behavior?
    \item In what ways are they deviating from normal behavior?
\end{itemize}

To achieve these goals, we use a \textbf{forecasting}-based approach, where we forecast the expected behavior of each sensor at each time based on the past. This allows the user to easily identify the sensors which deviate greatly from their expected behavior. Moreover, the user can compare the expected and observed behavior of each sensor, to understand why the model regards a sensor as anomalous. 

Thus, at time $t$, we define our model input $\mathbf{x}^{(t)} \in \mathbb{R}^{N \times w}$ based on a sliding window of size $w$ over the historical time series data (whether training or testing data): 
\begin{align}
    \mathbf{x}^{(t)} := \mathbf{\left[s^{(t-w)}, s^{(t-w+1)},\cdots,s^{(t-1)}\right]} 
\end{align}
The target output that our model needs to predict is the sensor data at the current time tick, i.e. $\mathbf{s^{(t)}}$.


\paragraph{Feature Extractor}
To capture the relationships between sensors, we introduce a graph attention-based feature extractor to fuse a node's information with its neighbors based on the learned graph structure. Unlike existing graph attention mechanisms, our feature extractor incorporates the sensor embedding vectors $\mathbf{v}_i$, which characterize the different behaviors of different types of sensors. To do this, we compute node $i$'s aggregated representation $\mathbf{z}_{i}$ as follows:
\begin{align}
    \mathbf{z}_{i}^{(t)} &= \mathsf{ReLU} \left( \alpha_{i, i} \mathbf{W} \mathbf{x}_{i}^{(t)}+\sum_{j \in \mathcal{N}(i)} \alpha_{i, j} \mathbf{W} \mathbf{x}_{j}^{(t)} \right),
\end{align}
where $\mathbf{x}_{i}^{(t)} \in \mathbb{R}^w $ is node i's input feature, $\mathcal{N}(i) =\{ {j \mid A_{ji} > 0} \}$ is the set of neighbors of node $i$ obtained from the learned adjacency matrix $A$, $\mathbf{W} \in \mathbb{R}^{d \times w}$ is a trainable weight matrix which applies a shared linear transformation to every node, and the attention coefficients $\alpha_{i,j}$ are computed as:
\begin{align}
    \mathbf{g}_{i}^{(t)} &= \mathbf{v}_{i} \oplus \mathbf{W} \mathbf{x}_{i}^{(t)} \label{eq:gi}\\
    \pi\left(i, j\right) &= \mathsf{LeakyReLU} \left(\mathbf{a}^{\top}\left( \mathbf{g}_{i}^{(t)}  \oplus \mathbf{g}_{j}^{(t)} \right)  \right) \\
    \alpha_{i, j} &= \frac{\exp \left(  \pi\left(i, j\right) \right) }{\sum_{k \in \mathcal{N}(i) \cup\{i\}} \exp \left(  \pi\left(i, k\right) \right)}, \label{eq:softmax}
\end{align}
where $\oplus$ denotes concatenation; thus $\mathbf{g}_{i}^{(t)}$ concatenates the sensor embedding $\mathbf{v}_{i}$ and the corresponding transformed feature $\mathbf{W} \mathbf{x}_{i}^{(t)}$, 
and $\mathbf{a}$ is a vector of learned coefficients for the attention mechanism. We use $\mathsf{LeakyReLU}$ as the non-linear activation to compute the attention coefficient, and normalize the attention coefficents using the softmax function in Eq. \eqref{eq:softmax}.

\paragraph{Output Layer}
From the above feature extractor, we obtain representations for all $N$ nodes, namely $\{\mathbf{z}_1^{(t)}, \cdots, \mathbf{z}_N^{(t)}\}$. For each $\mathbf{z}_i^{(t)}$, we element-wise multiply (denoted $\circ$) it with the corresponding time series embedding $\mathbf{v}_i$, and use the results across all nodes as the input of stacked fully-connected layers with output dimensionality $N$, to predict the vector of sensor values at time step $t$, i.e. $\mathbf{s^{(t)}}$:
\begin{align}
    \mathbf{\hat{s}^{(t)}} = f_\theta\left(\left[ \mathbf{v}_1 \circ \mathbf{z}_{1}^{(t)}, \cdots, \mathbf{v}_N \circ \mathbf{z}_{N}^{(t)}\right]\right)
\end{align}

The model's predicted output is denoted as $\mathbf{\hat{s}^{(t)}}$. We use the Mean Squared Error between the predicted output $\mathbf{\hat{s}^{(t)}}$ and the observed data, $\mathbf{s^{(t)}}$, as the loss function for minimization:
\begin{align}
    L_{\text{MSE}} = \frac{1}{\Ttr-w}\sum_{t=w+1}^{\Ttr}\left\|\mathbf{\hat{s}^{(t)}} - \mathbf{s^{(t)}}\right\|_{2}^{2}
\end{align}

\subsection{Graph Deviation Scoring}
Given the learned relationships, we want to detect and explain anomalies which deviate from these relationships. To do this, our model computes individual anomalousness scores for each sensor, and also combines them into a single anomalousness score for each time tick, thus allowing the user to localize which sensors are anomalous, as we will show in our experiments.

The anomalousness score compares the expected behavior at time $t$ to the observed behavior, computing an error value $\mathsf{Err}$ at time $t$ and sensor $i$:
\begin{align}
    \mathsf{Err}_i\left(t\right) &= |\mathbf{s_i^{(t)}} - \mathbf{\hat{s}_i^{(t)}} |
\end{align}
As different sensors can have very different characteristics, their deviation values may also have very different scales. To prevent the deviations arising from any one sensor from being overly dominant over the other sensors, we perform a robust normalization of the error values of each sensor:
\begin{align}
    a_i\left(t\right) &= \frac{ \mathsf{Err}_i\left(t\right) - \widetilde\mu_i }{\widetilde\sigma_i},
\end{align}
where $\widetilde\mu_i$ and ${\widetilde\sigma_i}$ are the median and inter-quartile range (IQR\footnote{IQR is defined as the difference between the 1st and 3rd quartiles of a distribution or set of values, and is a robust measure of the distribution's spread.}) across time ticks of the $\mathsf{Err}_i\left(t\right)$ values respectively. We use median and IQR instead of mean and standard deviation as they are more robust against anomalies. 

Then, to compute the overall anomalousness at time tick $t$, we aggregate over sensors using the max function (we use max as it is plausible for anomalies to affect only a small subset of sensors, or even a single sensor) :
\begin{align}
    A\left(t\right) &= \max_{i}a_i\left(t\right)
\end{align}

To dampen abrupt changes in values are often not perfectly predicted and result in sharp spikes in error values even when this behavior is normal, similar with ~\cite{hundman2018detecting}, we use a simple moving average(SMA) to generate the smoothed scores $A_{s}\left(t\right)$.

Finally, a time tick $t$ is labelled as an anomaly if $A_{s}(t)$ exceeds a fixed threshold. While different approaches could be employed to set the threshold such as extreme value theory~\cite{siffer2017anomaly}, to avoid introducing additional hyperparameters, we use in our experiments a simple approach of setting the threshold as the max of $A_{s}(t)$ over the validation data. 


\section{Experiments}
In this section, we conduct experiments to answer the following research questions:
\begin{itemize}
    \item \textbf{RQ1 (Accuracy):} Does our method outperform baseline methods in accuracy of anomaly detection in multivariate time series, based on ground truth labelled anomalies?
    \item \textbf{RQ2 (Ablation):} How do the various components of the method contribute to its performance?
    \item \textbf{RQ3 (Interpretability of Model):} How can we understand our model based on its embeddings and its learned graph structure?
    \item \textbf{RQ4 (Localizing Anomalies):} Can our method localize anomalies and help users to identify the affected sensors, as well as to understand how the anomaly deviates from the expected behavior?
\end{itemize}

\subsection{Datasets}
As real-world datasets with labeled ground-truth anomalies are scarce, especially for large-scale plants and factories, we use two sensor datasets based on water treatment physical test-bed systems: \texttt{SWaT} and \texttt{WADI}, where operators have simulated attack scenarios of real-world water treatment plants, recording these as the ground truth anomalies.

The Secure Water Treatment (\texttt{SWaT}) dataset comes from a water treatment test-bed coordinated by Singapore's Public Utility Board~\cite{dataset_swat}. It represents a small-scale version of a realistic modern Cyber-Physical system, integrating digital and physical elements to control and monitor system behaviors. As an extension of \texttt{SWaT}, Water Distribution (\texttt{WADI}) is a distribution system comprising a larger number of water distribution pipelines~\cite{dataset_wadi}. Thus \texttt{WADI} forms a more complete and realistic water treatment, storage and distribution network. The datasets contain two weeks of data from normal operations, which are used as training data for the respective models. A number of controlled, physical attacks are conducted at different intervals in the following days, which correspond to the anomalies in the test set.

Table \ref{table:dataset} summarises the statistics of the two datasets. In order to speed up training, the original data samples are down-sampled to one measurement every $10$ seconds by taking the median values. The resulting label is the most common label during the $10$ seconds. Since the systems took 5-6 hours to reach stabilization when first turned on  ~\cite{goh2016dataset}, we eliminate the first 2160 samples for both datasets.
\begin{table}[t]
    \centering
    \begin{tabular}{ cccccc } 
     \toprule
        \textbf{Datasets} & \textbf{\#Features} & \textbf{\#Train} & \textbf{\#Test} & \textbf{Anomalies} \\
     \midrule
        \texttt{SWaT} & 51 & 47515 & 44986 & 11.97\% \\
        \texttt{WADI} & 127 & 118795 & 17275 & 5.99\% \\
     \bottomrule
    \end{tabular}
    \caption{Statistics of the two datasets used in experiments}
    \label{table:dataset}
\end{table}
\subsection{Baselines}
We compare the performance of our proposed method with five popular anomaly detection methods, including:
\begin{itemize}
    \item \textbf{\textsc{PCA:}}  Principal Component Analysis ~\cite{pca} finds a low-dimensional projection that captures most of the variance in the data. The anomaly score is the reconstruction error of this projection.
    \item \textbf{\textsc{KNN:} } K Nearest Neighbors uses each point's distance to its $k$th nearest neighbor as an anomaly score~\cite{knn}.
    \item \textbf{\textsc{FB:}} A Feature Bagging detector is a meta-estimator that fits a number of detectors on various sub-samples of the dataset, then aggregates their scores~\cite{featurebagging}.
    \item \textbf{\textsc{AE:}} Autoencoders consist of an encoder and decoder which reconstruct data samples ~\cite{autoencoding}. It uses the reconstruction error as the anomaly score.
    \item \textbf{\textsc{DAGMM:}} Deep Autoencoding Gaussian Model joints deep Autoencoders and Gaussian Mixture Model to generate a low-dimensional representation and reconstruction error for each observation~\cite{dagmm}.
    \item \textbf{\textsc{LSTM-VAE:}} LSTM-VAE~\cite{lstmvae} replaces the feed-forward network in a VAE with LSTM to combine LSTM and VAE. It can measure reconstruction error with the anomaly score.
    \item \textbf{\textsc{MAD-GAN:}} A GAN model is trained on normal data, and the LSTM-RNN discriminator along with a reconstruction-based approach is used to compute anomaly scores for each sample~\cite{li2019mad}.
\end{itemize}

\subsection{Evaluation Metrics}
We use precision ($\mathsf{Prec}$), recall ($\mathsf{Rec}$) and F1-Score ($\mathsf{F1}$) over the test dataset and its ground truth values to evaluate the performance of our method and baseline models: $\mathsf{F1}=\frac{2 \times \mathsf{Prec} \times \mathsf{Rec}}{\mathsf{Prec} + \mathsf{Rec}}$, where $\mathsf{Prec}=\frac{\mathsf{TP}}{\mathsf{TP}+\mathsf{FP}}$ and $\mathsf{Rec}=\frac{\mathsf{TP}}{\mathsf{TP}+\mathsf{FN}}$, and $\mathsf{TP}, \mathsf{TN}, \mathsf{FP}, \mathsf{FN}$ are the numbers of true positives, true negatives, false positives, and false negatives. Note that our datasets are unbalanced, which justifies the choice of these metrics, which are suitable for unbalanced data. To detect anomalies, we use the maximum anomaly score over the validation dataset to set the threshold. 
At test time, any time step with an anomaly score over the threshold will be regarded as an anomaly. 

\subsection{Experimental Setup}
We implement our method and its variants in PyTorch~\cite{paszke2017automatic} version 1.5.1 with CUDA 10.2 and PyTorch Geometric Library~\cite{Fey/Lenssen/2019} version 1.5.0, 
and train them on a server with Intel(R) Xeon(R) CPU E5-2690 v4 @ 2.60GHz and 4 NVIDIA RTX 2080Ti graphics cards. The models are trained using the Adam optimizer with learning rate $ 1 \times 10^{-3} $ and $ (\beta_1, \beta_2)=(0.9,0.99) $. We train models for up to $50$ epochs and use early stopping with patience of $10$. We use embedding vectors with length of $128(64)$, $k$ with $30(15)$ and hidden layers of $128 (64)$ neurons for the \texttt{WADI} (\texttt{SWaT}) dataset, corresponding to their difference in input dimensionality. We set the sliding window size $w$ as $5$ for both datasets.

\begin{table}[t]
    \centering\fontsize{9}{11}\selectfont
    \begin{tabular}{rrrrrrr}
        \toprule
          & \multicolumn{3}{c}{\textbf{\texttt{SWaT}}} & \multicolumn{3}{c}{\textbf{\texttt{WADI}}} \\
         \cmidrule(r){2-4} \cmidrule(r){5-7}
         \textbf{Method} & $\mathbf{\mathsf{Prec}}$ & $\mathbf{\mathsf{Rec}}$ & $\mathbf{\mathsf{F1}}$ & $\mathbf{\mathsf{Prec}}$ & $\mathbf{\mathsf{Rec}}$ & $\mathbf{\mathsf{F1}}$ \\
         \midrule
         \textsc{PCA} & 24.92 & 21.63 & 0.23 & 39.53 & 5.63 & 0.10 \\
         \textsc{KNN} & 7.83 & 7.83 & 0.08 & 7.76 & 7.75 & 0.08 \\
         \textsc{FB}  & 10.17 & 10.17 & 0.10 & 8.60 & 8.60 & 0.09 \\
         \textsc{AE} & 72.63 & 52.63 & 0.61 & 34.35 & 34.35 & 0.34 \\
         \textsc{\footnotesize{DAGMM}} & 27.46 & 69.52 & 0.39 & 54.44 & 26.99 & 0.36 \\
         \textsc{\footnotesize{LSTM-VAE}} & 96.24 & 59.91 & 0.74 & 87.79 & 14.45 & 0.25 \\
         \textsc{\footnotesize{MAD-GAN}} & 98.97 & 63.74 & 0.77 & 41.44 & 33.92 & 0.37 \\
         \midrule
         \textsc{GDN} & \textbf{99.35} & \textbf{68.12} & \textbf{0.81} &\textbf{97.50} & \textbf{40.19} & \textbf{0.57} \\
         \bottomrule
    \end{tabular}
    \caption{Anomaly detection accuracy in terms of precision(\%), recall(\%), and F1-score, on two datasets with ground-truth labelled anomalies. Part of the results are from ~\cite{li2019mad}.}
    \label{tab:performance}
\end{table}

\subsection{RQ1. Accuracy}
In Table \ref{tab:performance}, we show the anomaly detection accuracy in terms of precision, recall and F1-score, of our \textsc{GDN} method and the baselines, on the \texttt{SWaT} and \texttt{WADI} datasets. The results show that \textsc{GDN} outperforms the baselines in both datasets, with high precision in both datasets of $0.99$ on \texttt{SWaT} and $0.98$ on \texttt{WADI}. In terms of F-measure, \textsc{GDN} outperforms the baselines on \texttt{SWaT}; on \texttt{WADI}, it has $54\%$ higher F-measure than the next best baseline. \texttt{WADI} is more unbalanced than \texttt{SWaT} and has higher dimensionality than \texttt{SWaT} as shown in Table \ref{table:dataset}. Thus, our method shows effectiveness even in unbalanced and high-dimensional attack scenarios, which are of high importance in real-world applications.



\subsection{RQ2. Ablation}
To study the necessity of each component of our method, we gradually exclude the components to observe how the model performance degrades. First, we study the importance of the learned graph by substituting it with a static complete graph, where each node is linked to all the other nodes. Second, to study the importance of the sensor embeddings, we use an attention mechanism without sensor embeddings: that is, $\mathbf{g}_{i} = \mathbf{W} \mathbf{x}_{i}$ in Eq. \eqref{eq:gi}. Finally, we disable the attention mechanism, instead aggregating using equal weights assigned to all neighbors. The results are summarized in Table \ref{tab:ablation} and provide the following findings:

\begin{itemize}
    \item Replacing the learned graph structure with a complete graph degrades performance in both datasets. The effect on the \texttt{WADI} dataset is more obvious. This indicates that the graph structure learner enhances performance, especially for large-scale datasets.
    \item The variant which removes the sensor embedding from the attention mechanism underperforms the original model in both datasets. This implies that the embedding feature improves the learning of weight coefficients in the graph attention mechanism.
    \item Removing the attention mechanism degrades the model's performance most in our experiments. Since sensors have very different behaviors, treating all neighbors equally introduces noise and misleads the model. This verifies the importance of the graph attention mechanism.
\end{itemize}
These findings suggest that \textsc{GDN}'s use of a learned graph structure, sensor embedding, and attention mechanisms all contribute to its accuracy, which provides an explanation for its better performance over the baseline methods.


\begin{table}[t]
    \centering
    \scalebox{0.9}{
    \begin{tabular}{lrrrrrr}
        \toprule
          & \multicolumn{3}{c}{\textbf{\texttt{SWaT}}} & \multicolumn{3}{c}{\textbf{\texttt{WADI}}} \\
         \cmidrule(r){2-4} \cmidrule(r){5-7}
         \textbf{Method} & $\mathbf{\mathsf{Prec}}$ & $\mathbf{\mathsf{Rec}}$ & $\mathbf{\mathsf{F1}}$ & $\mathbf{\mathsf{Prec}}$ & $\mathbf{\mathsf{Rec}}$ & $\mathbf{\mathsf{F1}}$ \\
         \midrule
         \textsc{GDN} & \textbf{99.35} & \textbf{68.12} & \textbf{0.81} &\textbf{97.50} & \textbf{40.19} & \textbf{0.57} \\
         \midrule
         - \textsc{TopK} & 97.41 & 64.70 & 0.78 & 92.21 & 35.12 & 0.51 \\
         \ \ \ - \textsc{Emb} & 92.31 & 61.25 & 0.76 & 91.86 & 33.49 & 0.49 \\
         \ \ \ \ \ \ - \textsc{Att} & 71.05 & 65.06 & 0.68 & 61.33 & 38.85 & 0.48 \\
         \bottomrule
    \end{tabular}}
    \caption{Anomaly detection accuracy in term of percision({\%}), recall({\%}), and F1-score of \textsc{GDN} and its variants.}
    \label{tab:ablation}
\end{table}

\begin{figure}[t]
\centering
\scalebox{0.9}{
\includegraphics[width=7cm]{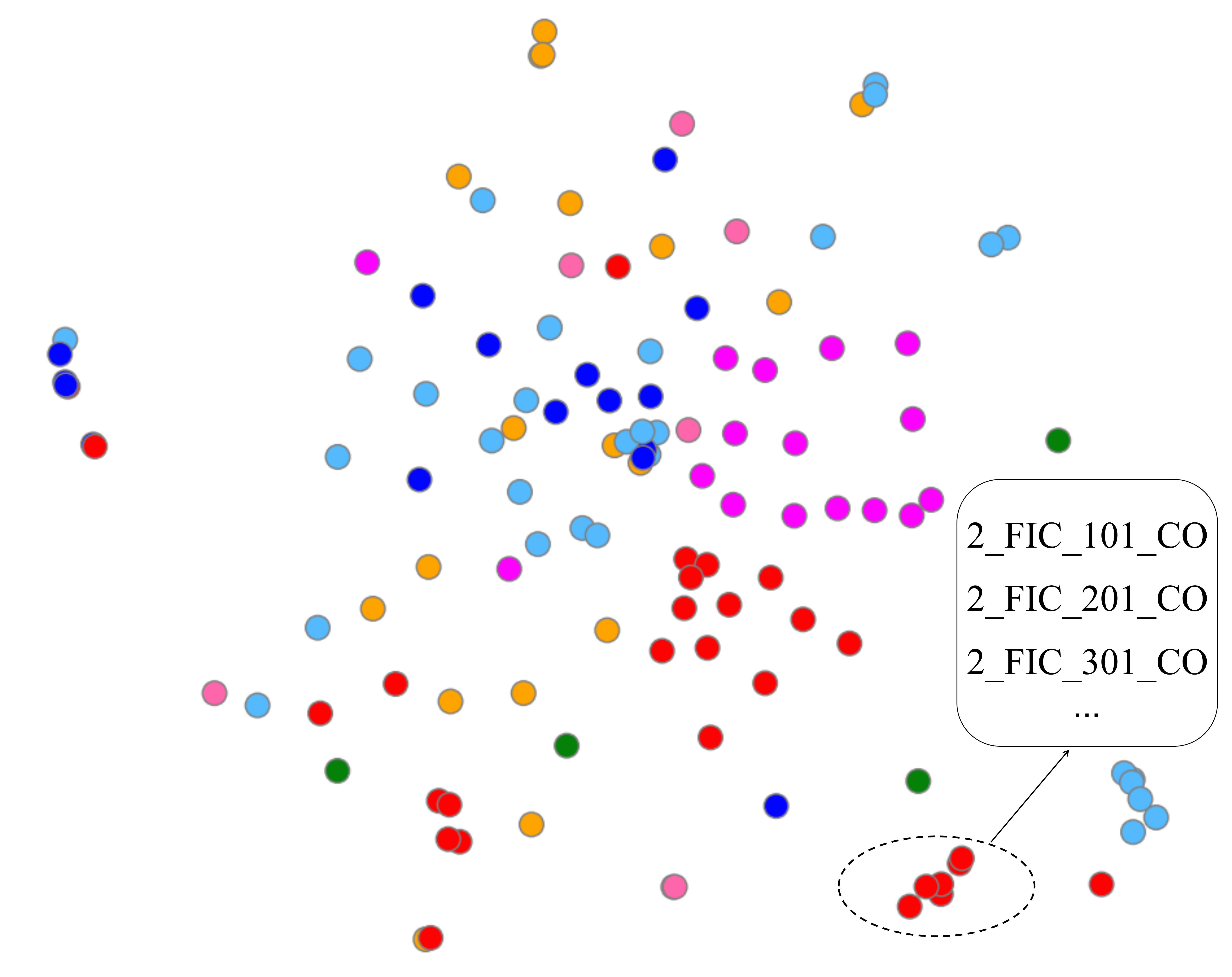}}
\caption{A t-SNE plot of the sensor embeddings of our trained model on the \texttt{WADI} dataset. Node colors denote classes. Specifically, the dashed circled region shows localized clustering of 2\_FIC\_x01\_CO sensors. These sensors are measuring similar indicators in \texttt{WADI}. 
}
\label{fig:feature_tsne}
\end{figure}

\begin{figure*}
\centering
\scalebox{0.85}{
\includegraphics[width=\textwidth]{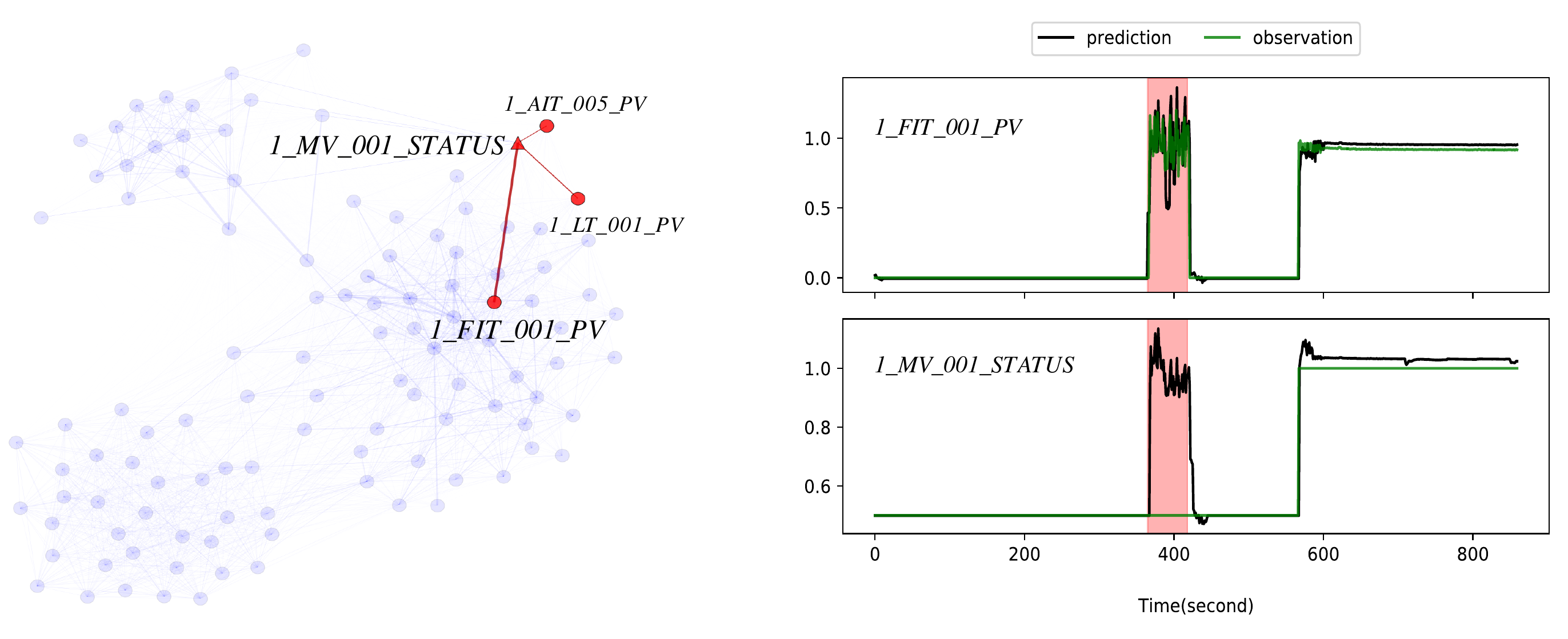}}
\caption{\emph{Left:} Force-directed graph layout with attention weights as edge weights, showing an attack in \texttt{WADI}. The red triangle denotes the central sensor identified by our approach, with highest anomaly score. Red circles indicate nodes with edge weights larger than $0.1$ to the central node. \emph{Right:} Comparing expected and observed data helps to explain the anomaly. The attack period is shaded in red.}
\label{fig:local}
\end{figure*}

\subsection{RQ3. Interpretability of Model}
\paragraph{Interpretability via Sensor Embeddings} To explain the learned model, we can visualize its sensor embedding vectors, e.g. using t-SNE\cite{tsne}, shown on the \texttt{WADI} dataset in Figure \ref{fig:feature_tsne}. Similarity in this embedding space indicate similarity between the sensors' behaviors, so inspecting this plot allows the user to deduce groups of sensors which behave in similar ways. 

To validate this, we color the nodes using $7$ colors corresponding to $7$ classes of sensors in \texttt{WADI} systems. The representation exhibits localized clustering in the projected 2D space, which verifies the effectiveness of the learned feature representations to reflect the localized sensors' behavior similarity. Moreover, we observe a group of sensors forming a localized cluster, shown in the dashed circled region. Inspecting the data, we find that these sensors measure similar indicators in water tanks that perform similar functions in the \texttt{WADI} water distribution network, explaining the similarity between these sensors.
\paragraph{Interpretability via Graph Edges and Attention Weights} Edges in our learned graph provide interpretability by indicating which sensors are related to one another. Moreover, the attention weights further indicate the importance of each of a node's neighbors in modelling the node's behavior. Figure \ref{fig:local} (left) shows an example of this learned graph on the \texttt{WADI} dataset. The following subsection further shows a case study of using this graph to localize and understand an anomaly. 

\subsection{RQ4. Localizing Anomalies}
How well can our model help users to localize and understand an anomaly? Figure \ref{fig:local} (left) shows the learned graph of sensors, with edges weighted by their attention weights, and plotted using a force-directed layout\cite{kobourov2012spring}. 

We conduct a case study involving an anomaly with a known cause: as recorded in the documentation of the \texttt{WADI} dataset, this anomaly arises from a flow sensor, \emph{1\_FIT\_001\_PV}, being attacked via false readings. These false readings are within the normal range of this sensor, so detecting this anomaly is nontrivial.

During this attack period, \textsc{GDN} identifies \emph{1\_MV\_001\_STATUS} as the deviating sensor with the highest anomaly score, as indicated by the red triangle in Figure \ref{fig:local} (left). The large deviation at this sensor indicates that \emph{1\_MV\_001\_STATUS} could be the attacked sensor, or closely related to the attacked sensor.


\textsc{GDN} indicates (in red circles) the sensors with highest attention weights to the deviating sensor. Indeed, these neighbors are closely related sensors: the \emph{1\_FIT\_001\_PV} neighbor is normally highly correlated with \emph{1\_MV\_001\_STATUS}, as the latter shows the valve status for a valve which controls the flow measured by the former. However, the attack caused a deviation from this relationship, as the attack gave false readings only to \emph{1\_FIT\_001\_PV}. \textsc{GDN} further allows understanding of this anomaly by comparing the predicted and observed sensor values in Figure \ref{fig:local} (right): for \emph{1\_MV\_001\_STATUS}, our model predicted an increase (as \emph{1\_FIT\_001\_PV} increased, and our model has learned that the sensors increase together). Due to the attack, however, no change was observed in \emph{1\_MV\_001\_STATUS}, leading to a large error which was detected as an anomaly by \textsc{GDN}. 

In summary: 1) our model's individual anomaly scores help to localize anomalies; 2) its attention weights help to find closely related sensors; 3) its predictions of expected behavior of each sensor allows us to understand how anomalies deviate from expectations.


\section{Conclusion}
In this work, we proposed our Graph Deviation Network (\textsc{GDN}) approach, which learns a graph of relationships between sensors, and detects deviations from these patterns, while incorporating sensor embeddings. Experiments on two real-world sensor datasets showed that \textsc{GDN} outperformed baselines in accuracy, provides an interpretable model, and helps users to localize and understand anomalies. Future work can consider additional architectures and online training methods, to further improve the practicality of the approach.

\section*{Acknowledgments}
This work was supported in part by NUS ODPRT Grant R252-000-A81-133. The datasets are provided by iTrust, Centre for Research in Cyber Security, Singapore University of Technology and Design.

\bibliography{reference}
 
\end{document}